\documentclass[letterpaper]{article} 
\PassOptionsToPackage{table}{xcolor} 

\usepackage[submission]{aaai2026}  
\usepackage{times}  
\usepackage{helvet}  
\usepackage{courier}  
\usepackage[hyphens]{url}  
\usepackage{graphicx} 
\def\UrlFont{\rm}  
\usepackage{natbib}  
\usepackage{caption} 
\usepackage{algorithm}
\usepackage{algorithmic}

\usepackage{multirow}
\usepackage{amsmath}
\usepackage{amsfonts}
\usepackage{booktabs}
\usepackage{arydshln}
\usepackage{subfigure}
\usepackage{hyperref}

\usepackage{newfloat}
\usepackage{listings}
\DeclareCaptionStyle{ruled}{labelfont=normalfont,labelsep=colon,strut=off} 
\floatstyle{ruled}
\newfloat{listing}{tb}{lst}{}
\floatname{listing}{Listing}
\title{2D Gaussian Splatting with Semantic Alignment for Image Inpainting}
\author{
    Hongyu Li\textsuperscript{\rm 1},
    Chaofeng Chen\textsuperscript{\rm 2},
    Xiaoming Li\textsuperscript{\rm 3},
    Guangming Lu\textsuperscript{\rm 1}
}
\affiliations{
    \textsuperscript{\rm 1}Harbin Institute of Technology, Shenzhen\\
    \textsuperscript{\rm 2}School of Artificial Intelligence, Wuhan University\\
    \textsuperscript{\rm 3}Nanyang Technological University\\

    \href{https://github.com/hitlhy715/2DGS_inpaint}{https://github.com/hitlhy715/2DGS-inpaint}

}

\begin{document}

\maketitle

\begin{abstract}
Gaussian Splatting (GS), a recent technique for converting discrete points into continuous spatial representations, has shown promising results in 3D scene modeling and 2D image super-resolution. In this paper, we explore its untapped potential for image inpainting, which demands both locally coherent pixel synthesis and globally consistent semantic restoration. We propose the first image inpainting framework based on 2D Gaussian Splatting, which encodes incomplete images into a continuous field of 2D Gaussian splat coefficients and reconstructs the final image via a differentiable rasterization process. The continuous rendering paradigm of GS inherently promotes pixel-level coherence in the inpainted results. To improve efficiency and scalability, we introduce a patch-wise rasterization strategy that reduces memory overhead and accelerates inference. For global semantic consistency, we incorporate features from a pretrained DINO model. We observe that DINO’s global features are naturally robust to small missing regions and can be effectively adapted to guide semantic alignment in large-mask scenarios, ensuring that the inpainted content remains contextually consistent with the surrounding scene.
Extensive experiments on standard benchmarks demonstrate that our method achieves competitive performance in both quantitative metrics and perceptual quality, establishing a new direction for applying Gaussian Splatting to 2D image processing.

\end{abstract}


\section{Introduction}

Human visual perception naturally interprets the world as a continuous experience. In contrast, digital images are constrained by hardware and storage limitations, resulting in representations composed of discrete pixels. Similarly, prevailing neural network architectures, such as Convolutional Neural Networks (CNNs) and Transformers, operate on pixel-based inputs and rely on fixed spatial encodings. This fundamental mismatch limits the ability of networks to fully capture the continuous nature of real-world visual data, particularly in the spatial domain, thereby restricting their expressive capacity and representational fidelity. 

To address this challenge, recent research has increasingly focused on integrating the complementary strengths of discrete and continuous representations~\cite{tian2023addp, jiao2025unitoken, jiang2024maven}. Implicit Neural Representations (INRs) \cite{chen2021learning, cao2023ciaosr, wu2023learning, li2022adaptive} have made significant strides by learning mappings from spatial coordinates to RGB values, enabling finer spatial granularity and continuous modeling. Latest methods \cite{peng2025pixel, hu2025Gaussiansr} employ 2D Gaussian Splatting to naturally encode local continuous features at arbitrary scales.

\begin{figure}
    \centering
    \includegraphics[width=1\linewidth]{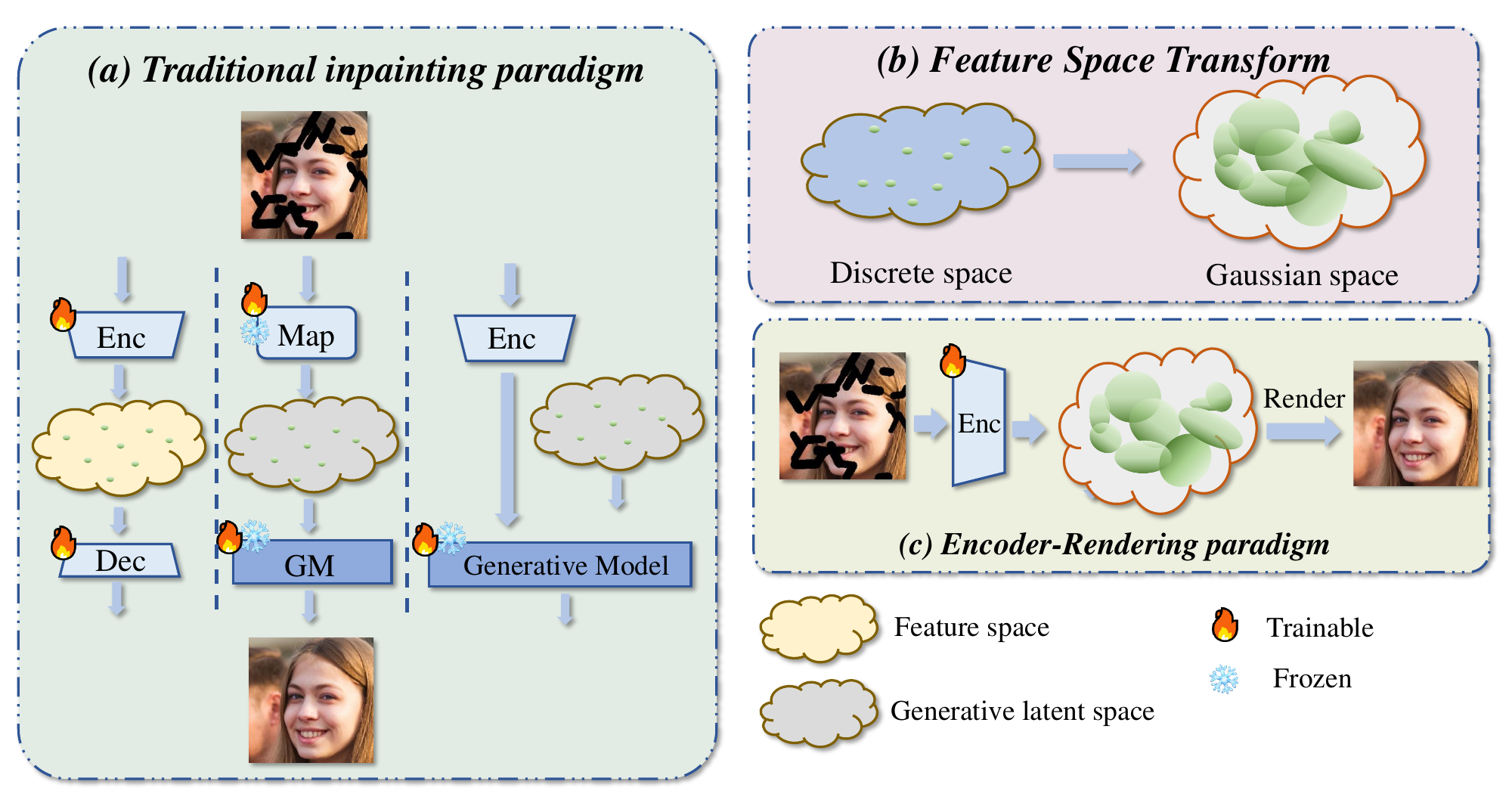}
    \caption{(a) Traditional inpainting methods rely on CNNs or Transformers to synthesize missing pixels in a spatially discrete manner. (b) Gaussian splatting generates pixels in a continuous way. (c) Our method introduces a novel paradigm: instead of synthesizing pixel values directly, we encode the input image into a learned Gaussian feature space and reconstruct it through a differentiable rasterization process, allowing for continuous and smoother pixel generation.}
    \label{fig_moti}
    \vspace{-1em}
\end{figure}

Meanwhile, the potential of Gaussian Splatting (GS) for image inpainting remains underexplored. GS represents images as continuous fields using localized, overlapping Gaussians, enabling smooth interpolation and fine-grained detail reconstruction. These properties make it well-suited for filling in missing regions with both spatial continuity and semantic coherence. In contrast, as illustrated in Figure~\ref{fig_moti}(a), most existing inpainting methods rely on CNNs or Transformers to synthesize missing pixels. However, their inherently discrete spatial processing often hinders the reconstruction of coherent pixel-level structures, particularly in regions with complex textures or fine details.



In this paper, we introduce a novel approach to image inpainting based on Gaussian splatting, positing the hypothesis that a complete Gaussian representation of an image can be modeled from its incomplete regions, leveraging the inherent continuity of Gaussian functions to restore high-quality images. Our framework maps incomplete images to dense Gaussian feature fields, which are then rendered back to the pixel domain to produce complete images, as illustrated in Figure \ref{fig_moti}(c).

To address the computational challenges of high-resolution inpainting, we introduce a patch-level rasterization strategy. This approach processes images in smaller, manageable segments, significantly reducing GPU memory demands and computational overhead by exploiting the local coherence inherent in image patches. While this patch-based processing improves efficiency, it introduces challenges in maintaining global semantic consistency across patches.

To ensure global semantic coherence, we investigate the robustness of DINO features for inpainting tasks. We find that DINO features demonstrate remarkable robustness to small masks and can be effectively adapted to handle large masks through simple adaptation techniques. Building on this observation, we propose incorporating DINO features as global semantic guidance within our Gaussian splatting framework. This integration enables the model to maintain semantic consistency across patches while preserving the computational benefits of our patch-level approach.

The main contributions of this paper are:
\begin{itemize}
    \item \textbf{A novel 2D Gaussian Splatting framework for image inpainting.} To the best of our knowledge, this is the first approach to leverage 2D Gaussian splatting for high-quality image inpainting. Our method directly maps known image pixels into a continuous Gaussian feature space, enabling effective reconstruction of missing regions through the inherent continuity of Gaussian functions without requiring explicit optimization of scene-wide parameters from scratch.

    \item \textbf{Patch-level rasterization for scalable high-resolution processing.} We introduce a patch-wise processing strategy that addresses the computational challenges of high-resolution inpainting by dividing images into manageable segments. This approach significantly reduces GPU memory consumption and accelerates rendering while incorporating overlapping regions and blending techniques to maintain spatial continuity across patch boundaries.
    \item \textbf{DINO feature adaptation for global semantic guidance.} We demonstrate that DINO features exhibit remarkable robustness to small masks and can be effectively adapted to handle large masks through simple techniques. Leveraging this discovery, we integrate DINO features as global semantic guidance via Adaptive Layer Normalization (AdaLN), enabling the model to maintain semantic consistency across patches while conditioning representations on high-level contextual cues for accurate reconstruction of complex structures and object relationships.
\end{itemize}





\section{Related Work}

\subsection{Gaussian Splatting}
\paragraph{3D Gaussian Splatting}
3D Gaussian Splatting (3DGS) \cite{kerbl20233d} has recently emerged as a compelling alternative paradigm in computer graphics, enabling 3D scene synthesis via explicit, learnable Gaussian representations. Unlike implicit neural representations (INRs) such as NeRF~\cite{mildenhall2021nerf}, which rely on coordinate-based function mappings, 3DGS represents scenes using a large number of parameterized Gaussians combined with differentiable rasterization techniques, supporting real-time rendering and fine-grained editability. This approach has shown remarkable effectiveness in various applications \cite{keetha2024splatam,luiten2024dynamic,chen2024text,huang2024sc}.

\paragraph{2D Gaussian Splatting}
Building upon 3DGS, 2D Gaussian Splatting (2DGS) exploits similar principles in image processing, benefiting from favorable mathematical properties for representing continuous image information. Recently, 2DGS has demonstrated success in tasks such as image tokenization and super-resolution. Methodologically, existing works fall into two categories: 1) Using Gaussians to diffuse or broadcast image features. For example, GaussianSR \cite{hu2025Gaussiansr} constructs a continuous feature space for Arbitrary-Scale Super-Resolution (ASSR) by diffusing pixel-level features into a higher-resolution space. Similarly, GaussianToken \cite{dong2025Gaussiantoken} uses 2DGS to overcome the limited representation capacity caused by discrete feature spaces in vector quantization methods \cite{esser2021taming}. 2) Direct modeling of Gaussian spaces. GaussianImage \cite{zhang2024Gaussianimage} pioneers the use of 2DGS for image compression by optimizing Gaussian parameters, although their method currently processes only one image at a time. GSASR \cite{chen2025generalized} proposes a feature injection module that constructs a 2D Gaussian space via learnable embeddings. PixeltoGaussian \cite{peng2025pixel} empirically analyzes Gaussian space properties, deriving priors that improve early-stage convergence speed and training stability. To the best of our knowledge, however, the potential of 2D Gaussian Splatting for image inpainting remains unexplored.

\begin{figure*}[htbp]
\centering
\includegraphics[width=0.98\textwidth]{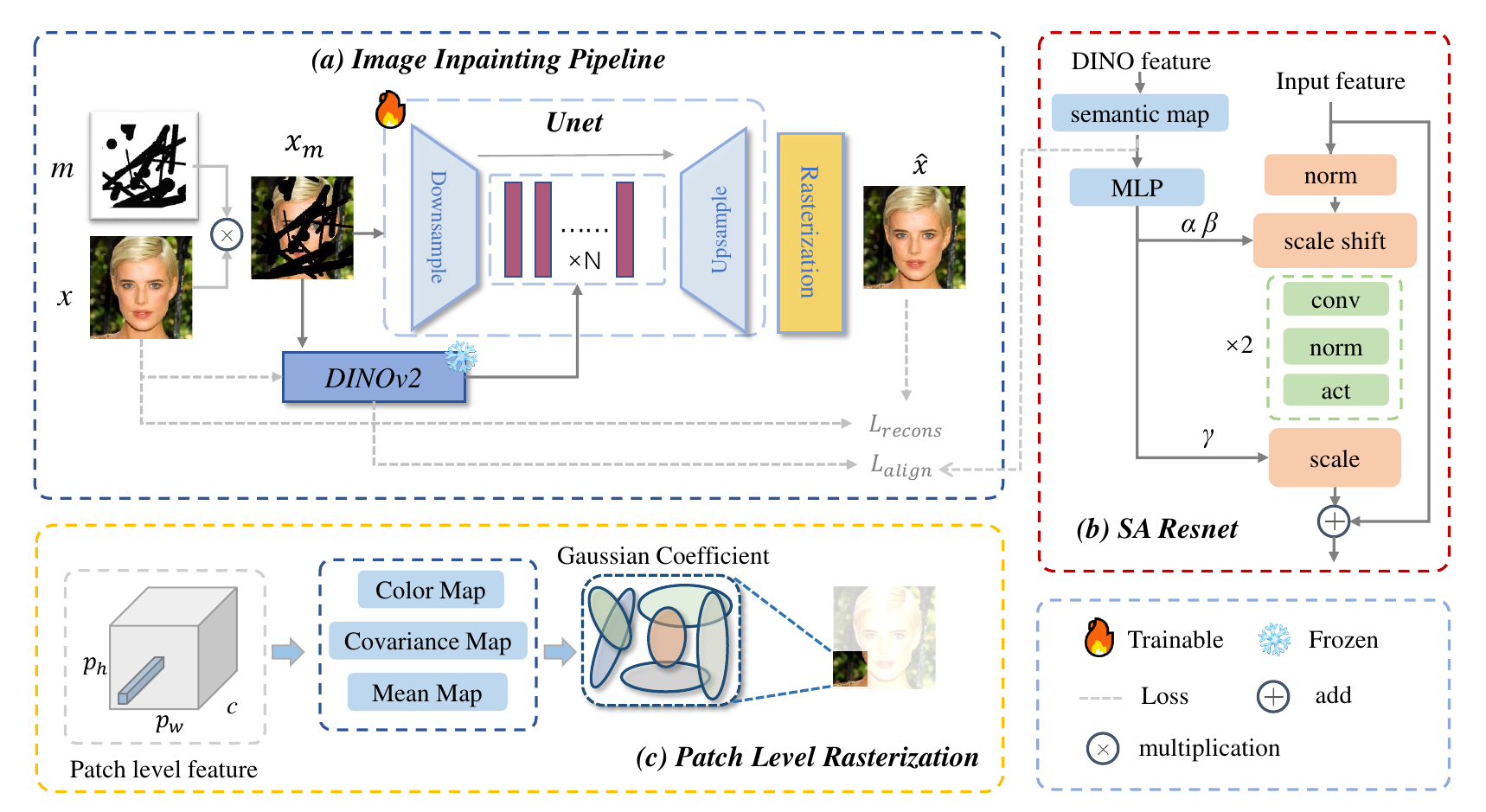} 
\caption{{Overview of our proposed framework:} (a) The overall pipeline consists of a U-Net architecture, DINO-based semantic alignment, and a differentiable rasterization module. (b) The Semantic Alignment (SA) ResNet integrates high-level semantic priors from DINO using an AdaLN-based modulation. (c) The Patch-Level Rasterization module transforms patch-wise Gaussian parameters into continuous representations, which are then composited to generate the complete image.}
\label{fig-overall}
\vspace{-1em}
\end{figure*}

\subsection{Image Inpainting}

Image inpainting is a long-standing research problem that has undergone substantial evolution from early exemplar-based methods~\cite{efros1999texture,he2014image} to modern learning-based approaches~\cite{pathak2016context,liu2018image,yu2018generative,guo2021image}. The introduction of perceptual loss~\cite{johnson2016perceptual} and adversarial training~\cite{goodfellow2014generative} has further propelled the field by enabling models to generate visually plausible and semantically coherent content.

Convolutional neural networks (CNNs) have been widely adopted for inpainting \cite{liu2018image, yu2019free, zheng2019pluralistic}. While effective, CNN-based models are fundamentally limited by their restricted receptive fields. To address this, recent approaches have turned to transformer architectures for their superior capacity to model long-range dependencies~\cite{li2022mat,chang2022maskgit,deng2022t,liu2023coordfill}. However, despite their expressive power, transformers may still face challenges in capturing local image structures or handling high-resolution inputs efficiently due to architectural and computational constraints.

The rise of diffusion models, particularly Stable Diffusion~\cite{rombach2022high}, has enabled strong generative priors for inpainting. Methods~\cite{xie2023smartbrush, ju2024brushnet, deng2025acquire} project corrupted images into latent spaces for restoration, while others~\cite{yu2023inpaint, wasserman2024paint, tianyidan2025anywhere} incorporate LLMs for controllable, semantically guided inpainting. Despite their quality, these models often suffer from slow inference and high computational costs.

Recent attempts to integrate Gaussian Splatting with diffusion models~\cite{feinashley2024diffusionmodelsanisotropicGaussian} show limited performance on small, outdated datasets. In contrast, our work pioneers a dedicated 2D Gaussian Splatting framework for inpainting, offering a lightweight, semantically aware framework.

\section{Method}
The image inpainting task aims to reconstruct a complete image from a partial image with reasonable content, requiring overall coherence and clarity in the reconstructed result. In this section, we present our end-to-end encoder–rasterization framework with semantic alignment for image inpainting. We first review the basis of Gaussian splatting, then describe our complete inpainting pipeline with an efficient patch-level rasterization strategy with overlap smoothing. Finally, we present our semantic alignment strategy, which leverages high-level priors to guide feature adaptation and improve global consistency.

\subsection{Preliminary: Gaussian Splatting}
\label{3_1}
Gaussian Splatting (GS)~\cite{kerbl20233d} has demonstrated remarkable capabilities in the field of 3D view synthesis and is naturally suited for visual representation tasks due to its hybrid discrete-continuous nature. The image is represented by numerous 2D Gaussians, each Gaussian characterized by mean $\mu_i \in \mathbb{R}^2$, covariance matrix $\Sigma_i \in \mathbb{R}^{2 \times 2}$, coefficient $\sigma_i \in \mathbb{R}^1$, and color $c_i \in \mathbb{R}^3$.

Typically, the covariance matrix of a Gaussian should be positive semi-definite. Therefore, inspired by~\cite{zhang2024Gaussianimage}, $\Sigma_i$ can be factorized into the product of a lower triangular matrix $L_i \in \mathbb{R}^{2 \times 2}$ and its conjugate transpose $L_i^T$, which ensures positive semi-definiteness and reduces the covariance representation from 4 to 3. We denote this parameter set as the Gaussian space $\Theta$.
\begin{equation}
    \Sigma_i = L_iL_i^T \quad \Theta = \{ \mu, L, c, \tilde{\sigma} \}
\end{equation}

Primarily due to the favorable mathematical properties of the Gaussian Mixture Model(GMM)~\cite{reynolds2009Gaussian}, complex distributions can be constructed by multiple Gaussian kernels. Given the Gaussian field at position $(x,y)$, we represent the pixel coordinate as a 2D vector $\mathbf{p} = [x, y]^T$. The pixel value is computed as the sum of Gaussians:
\begin{equation}
I_{\mathbf{p}} = \sum_i c_i \sigma_i \exp\left(-\frac{1}{2}(\mathbf{p} - \mu_i)^T \Sigma_i^{-1} (\mathbf{p} - \mu_i)\right)
\label{eq:gauss_render}
\end{equation}

Therefore, given enough Gaussians, the image can be generated in the above way, where the relevant parameters of the Gaussians can generally be trainable.




\subsection{2D Gaussian Splatting Framework}
\label{3-2}


Building upon the above formulations, our pipeline consists of two main stages: (1) Gaussian feature encoding, where the input image is projected into a Gaussian parameter space; and (2) Rasterization-based rendering, which reconstructs the complete image by rendering the learned Gaussian fields. A conceptual overview is illustrated in Figure~\ref{fig-overall}(a).

\paragraph{Gaussian feature encoding}
Traditional methods often rely on latent feature embeddings obtained from neural networks, followed by a trainable decoder that maps these features back to pixel space. However, this approach typically relies on the training performance of the decoder and struggles to explicitly implement pixel-level continuity naturally. Therefore, our method models the image feature as the 2D Gaussian features, which are concrete in Gaussian space and continuous in each Gaussian domain. Each Gaussian kernel defines a smooth and differentiable field over the image plane, allowing the model to naturally propagate visual information from observed to missing regions. This spatially overlapping and continuous formulation is fundamentally advantageous for achieving pixel-level continuity.

Formally, given a masked input image $I_{\text{mask}} \in \mathbb{R}^{3\times H\times W}$, we employ a Unet encoder to extract a set of $N$ Gaussian features.  The encoder outputs a dense feature map $F_g \in \mathbb{R}^{C'\times H\times W}$ which is then downsampled via a strided convolution layer to a Gaussian-level feature $F_g^{'} \in \mathbb{R}^{N\times C'}$, where $N$ denotes the number of Gaussian kernels.

The encoder’s hierarchical structure, combined with skip connection, captures rich context while enhancing training stability. Moreover, the Gaussian mean $\mu$, which has been empirically observed to be sensitive in the training process, is initialized uniformly across the 2D plane as $\mu_{fix}$ and only the mean offset $\mu_{bias}$ is learned. 

\begin{figure*}
    
    \centering
    \includegraphics[width=1\linewidth]{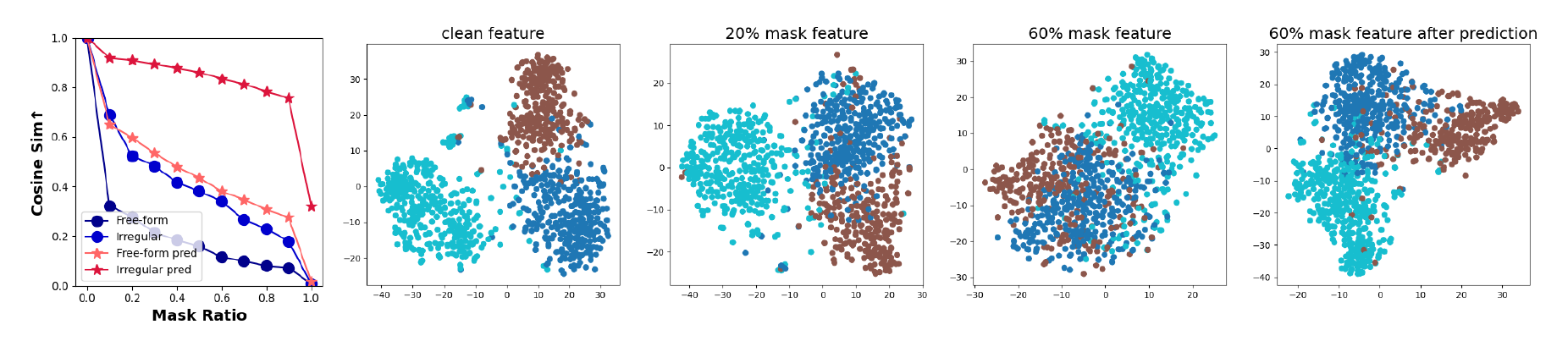}
    \vspace{-2em}
    \caption{{Representation Gap with Mask Image}: (a) Cosine similarity between masked and unmasked DINOv2 features drops with higher mask ratios, while the mapping block preserves similarity. (b) t-SNE plots show improved cluster separation and compactness with the block, indicating enhanced semantic consistency and reduced representation degradation under masking.}
    
    \label{fig-feature}
    \vspace{-1em}
\end{figure*}

\paragraph{Parameter Decoding}
Instead of a neural network decoder outputting discrete pixel values, we decode Gaussian parameters from $F_g^{'}$ using a set of lightweight multilayer perceptrons (MLPs):
\begin{equation}
    \mu_{bias} = E_\mu(F_g'), \quad c=E_c(F_g'), \quad l=E_l(F_g')
\end{equation}
where $\mu_{\text{bias}} \in \mathbb{R}^{N \times 2}$ denotes the learnable positional offsets, $c \in \mathbb{R}^{N \times 3}$ and $l \in \mathbb{R}^{N \times 3}$ represent the color and covariance parameters. To constrain the spatial extent of $\mu_{\text{bias}}$, we apply a \texttt{tanh} activation, restricting its values to the range $[-1, 1]$. The final positions are obtained by adding these learnable offsets to uniformly initialized positions: $\mu = \mu_{\text{bias}} + \mu_{\text{fix}}$.

These parameters are subsequently rasterized onto the image plane as described in Section 3.1, and their soft overlapping nature guarantees smooth color and intensity transitions across pixels, ensuring seamless image inpainting.

\subsection{Patch-level Rasterization}

However, achieving high-fidelity image representation with Gaussian kernels typically requires a large number of Gaussians, which scales with image resolution and leads to significant GPU memory consumption during rasterization. Moreover, directly generating Gaussian parameters from high-dimensional latent features for the entire image results in a large parameter set, further exacerbating memory pressure and slowing down rendering.

To mitigate this, we leverage the spatial locality of natural images and propose a patch-level 2D rasterization approach. Instead of managing a single global Gaussian set, we divide the image into multiple non-overlapping patches and assign a dedicated set of Gaussians to each.

Formally, an image $I \in \mathbb{R}^{H\times W}$ is conceptually divided into a grid of $N_p = \frac{H}{p} \times \frac{W}{p}$ non-overlapping patches, each of nominal size $p\times p$. For each patch $(i, j)$ corresponding to grid coordinates, we maintain a dedicated set of $N_{patch}$ Gaussian kernels. 
Then, parameter space $\Theta$ transforms into patch level space:
\begin{equation}
    \mu \in \mathbb{R}^{N_p\times N\times 2}, \quad c \in \mathbb{R}^{N_p\times N\times 3}, \quad l\in \mathbb{R}^{N_p\times N\times 3}
\end{equation}

While it might not reduce the total number of Gaussians ($N_p\times N_{patch}$) required for equivalent quality compared to a global approach, it reduces the number of Gaussians that need to be loaded and processed simultaneously during the rasterization of any single patch. This reduction in concurrent memory demand alleviates GPU memory pressure. Furthermore, since each patch can be rendered independently, the rasterization process is highly parallelizable across patches, leading to faster overall rendering times.

An inherent challenge with processing images in independent patches is the visible discontinuities at the patch boundaries, as the representation of one patch is independent of its neighbors. To ensure smoothness across patch borders, we employ a patch overlap strategy: during rendering, each patch is processed not just over its nominal $p\times p$ area, but over an extended region that includes a border of $a$ pixels on all four sides. This means each patch $(i,j)$ is rasterized over an area of size $(p+2a) \times (p+2a)$, using its parameter set $\Theta_{i,j}$, to produce a rendered patch $R_{i,j}$. The rendered patches $R_{i,j}$ thus overlap with their neighbors. 
To construct the final image, we retain the central $(p-2a) \times (p-2a)$ region of each $R_{i,j}$ and blend the overlapping border areas with neighboring patches. Blending is typically done using weighted averages based on pixel distance from the patch center or boundary, ensuring smooth transitions.

\subsection{Feature Adaptation For Semantic Guidance}

We hypothesize that semantic priors from pretrained models can provide valuable global context for image inpainting, enabling more cohesive and semantically faithful completion. To this end, we adopt features extracted from the pretrained DINOv2 model~\cite{oquab2023dinov2} as guidance signals within our framework.

A critical challenge arises from the fact that the input to the model is a masked image, and it is unclear whether features extracted from such incomplete inputs are meaningful or reliable. To investigate this, we conduct experiments illustrated in Figure~\ref{fig-feature}, which shows that DINO features extracted from lightly masked images still retain rich semantic information and can effectively guide inpainting. However, as the mask size increases, the extracted features become increasingly corrupted and less informative. This degradation limits their effectiveness in directly guiding the inpainting process for large missing regions.

To address this, we propose a simple yet effective feature adaptation module that transforms noisy masked features into semantically coherent representations. Specifically, we employ a lightweight MLP to learn a mapping from the masked feature space to an estimated clean feature space. As demonstrated in Figure~\ref{fig-feature}, this transformation significantly improves feature quality and enables robust semantic guidance across a range of mask sizes. The adapted features thus serve as conditional inputs to the inpainting network.

For integrating the adapted semantic features into the inpainting process, we adopt the Adaptive Layer Normalization (AdaLN) mechanism~\cite{karras2019style}, which is both parameter-efficient and capable of modulating network activations globally.

As shown in Figure~\ref{fig-overall}, given a predicted semantic feature vector $f_{\text{pred}} \in \mathbb{R}^{C_0}$ and a hidden feature map $f \in \mathbb{R}^{C\times H \times W}$, the AdaLN operation is defined as:
\begin{equation}
    AdaLN(f_{pred}, f) = B(LN(x)\times \alpha_{pred}+\beta_{pred})\times \gamma_{pred}
\end{equation}
where $B(\cdot)$ denotes the main processing block, and $\alpha_{\text{pred}}, \beta_{\text{pred}}, \gamma_{\text{pred}} \in \mathbb{R}^d$ are learned affine parameters obtained via linear projection from $f_{\text{pred}}$.

To explicitly align the predicted features with the ground truth semantic representations, we use a feature alignment loss based on negative cosine similarity:
\begin{equation}
\mathcal{L}_{\text{align}} = - \frac{f_{\text{clean}} \cdot f_{\text{pred}}}{|f_{\text{clean}}| \cdot |f_{\text{pred}}|}
\end{equation}
which encourages the predicted features to align with the direction of the clean features, thereby enhancing global semantic consistency during inpainting.

\begin{figure*}
    \vspace{-2em}
    \centering
    \includegraphics[width=1\linewidth]{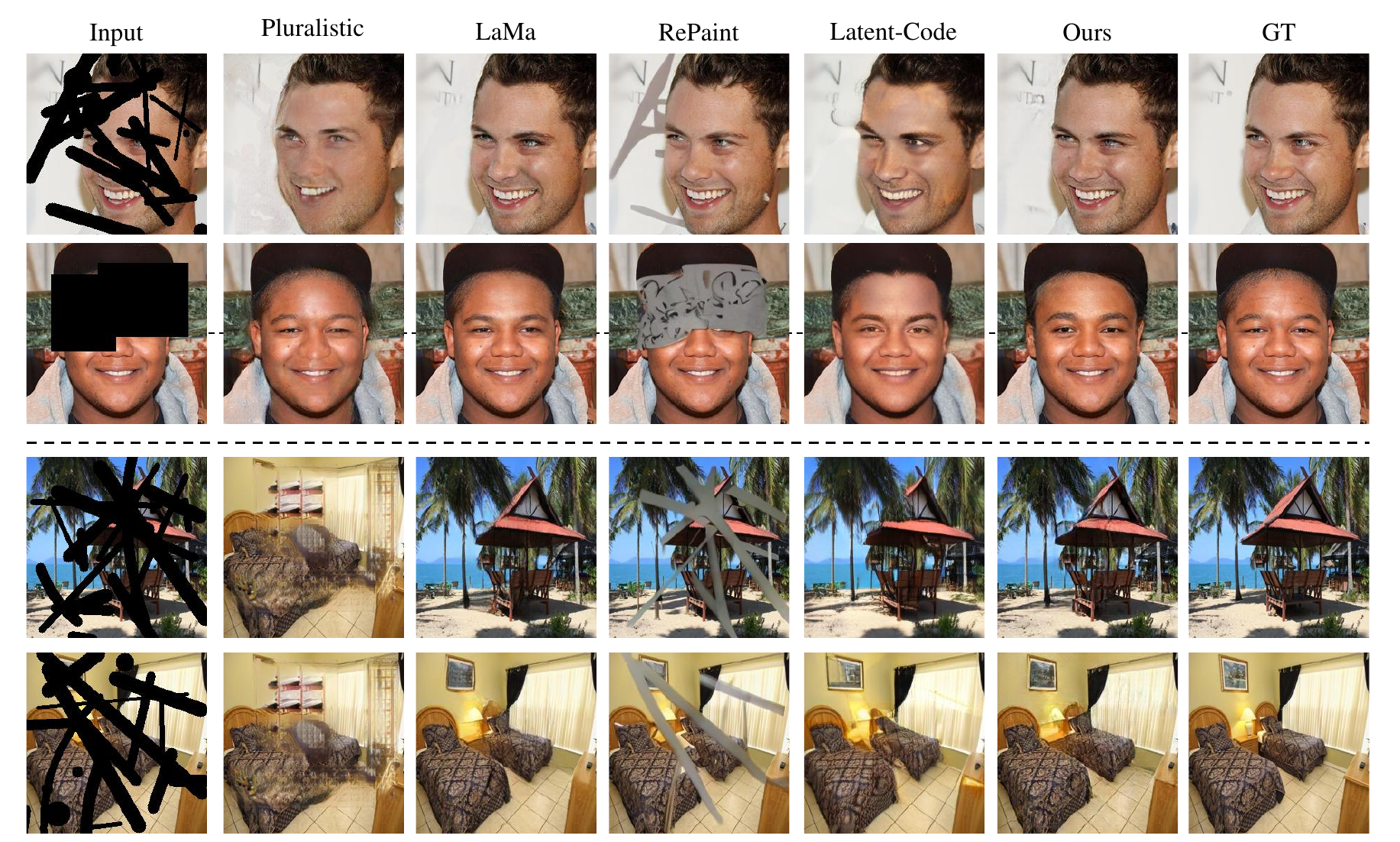}
    \caption{{Qualitative Results}. The top two rows show face inpainting results from the CelebA-HQ dataset, while the bottom two rows display natural scenes from the Places2 dataset with either irregular or regular patterns.}
    \label{fig-vis}
    \vspace{-1em}
\end{figure*}

\section{Experiments}
\subsection{Implementation Detail}
\paragraph{Architecture}
We adopt a simple convolutional U-Net as the image encoder, consisting of 3 downsampling layers, 9 bottleneck (middle) layers, and 3 upsampling layers. For semantic feature alignment, we utilize ViT-14 pretrained weights from DINOv2 and inject the extracted features into all 9 bottleneck layers of the encoder. To enhance training stability and accelerate convergence, especially during the early stages, we incorporate the Gaussian prior initialization strategy introduced in~\cite{peng2025pixel}. The network takes as input only the masked images, without explicitly providing the binary mask itself.

\paragraph{Loss Function}
The model is optimized using a composite loss function that balances reconstruction accuracy, perceptual quality, adversarial realism, and semantic alignment. The total objective is defined as:
\begin{equation}
\mathcal{L}_{\text{total}} = w_1 \mathcal{L}_{\text{recons}} + w_2 \mathcal{L}_{\text{gan}} + w_3 \mathcal{L}_{\text{lpips}} + w_4 \mathcal{L}_{\text{align}}
\end{equation}
where the weights are empirically set as $w_1:w_2:w_3:w_4 = 1:0.3:3:1$. The $\mathcal{L}_{recons}$ loss measures the absolute difference between the rendered pyramid and the ground truth. The $\mathcal{L}_{gan}$ is an adversarial loss employing a discriminator, similar to those used in SD~\cite{rombach2022high}, trained to distinguish between images rendered from the predicted Gaussians and real images. The $\mathcal{L}_{lpips}$ loss captures perceptual similarity based on deep features. The $\mathcal{L}_{align}$ enforces semantic consistency by minimizing the discrepancy between the DINO features of the masked input and those of the clean image.

\paragraph{Datasets and Metrics}
We conduct experiments on the two most commonly used datasets: Celeba-HQ~\cite{karras2017progressive} and Places2~\cite{zhou2017places}. Celeba-HQ contains 30,000 high-quality face images. We use 28,000 images for training and 2,000 images for evaluation. The train-test split follows LaMa~\cite{suvorov2022resolution}. We choose Places365, which has 1.8M natural images for training and commonly used 36,500 validation images for evaluation. To test the inpainting results, we choose the following metrics: FID~\cite{heusel2017gans}, LPIPS~\cite{zhang2018unreasonable}.

\paragraph{Training Details}
All models are trained with a batch size of 64 using the Adam optimizer and an initial learning rate of $2 \times 10^{-4}$. The default patch size is set to $16 \times 16$, and each Gaussian element is represented with a 12-dimensional hidden embedding. All experiments are conducted on 8 NVIDIA A800 GPUs.

\subsection{Qualitative Comparison}
Figure~\ref{fig-vis} presents qualitative comparisons across various image inpainting methods on two datasets: CelebA-HQ (top two rows) and Places2 (bottom two rows). It is evident that some methods, such as Latent-Code, RePaint, and Pluralistic, exhibit noticeable artifacts or semantic inconsistencies, particularly in complex textures and facial features. In contrast, our method produces visually coherent and semantically plausible completions.

\subsection{Quantitative Comparison}
To evaluate our method, we compare against representative baselines: Latent-Code~\cite{chen2024don}, Pluralistic~\cite{zheng2019pluralistic}, ZITS++~\cite{cao2023zits++}, RePaint~\cite{lugmayr2022repaint}, LaMa~\cite{suvorov2022resolution}, MAT~\cite{li2022mat}, as summarized in Table~\ref{tab-celeba}. Experiments are conducted on both regular and irregular masks of varying sizes, using publicly available checkpoints and identical image-mask pairs for fair comparison.

Our method matches or outperforms state-of-the-art baselines in both fidelity and perceptual quality. It preserves identity well on face datasets and generalizes effectively to complex natural scenes. 
While our FID is slightly higher than LaMa’s, this may result from its frequency-domain modeling and multi-scale fusion, which better capture global structure. Our approach, by contrast, focuses on local continuity and semantic coherence. As FID emphasizes distribution alignment, it may slightly favor methods like LaMa.

\begin{table*}[htbp]
    \renewcommand\arraystretch{1.0}
    \centering
    \small
    \begin{tabular}{l|cccccc|cccc}
        \hline
        \multirow{3}{*}{Method} & \multicolumn{6}{c|}{\textbf{Celeba-HQ}} & \multicolumn{4}{c}{\textbf{Places2}} \\
        \cline{2-11}
        & \multicolumn{2}{c|}{\textbf{Small}} & \multicolumn{2}{c|}{\textbf{Large}} & \multicolumn{2}{c|}{\textbf{Regular}} & \multicolumn{2}{c|}{\textbf{Small}} & \multicolumn{2}{c}{\textbf{Large}} \\
         
        & FID↓ & LPIPS↓  &  \multicolumn{1}{|c}{FID↓} & LPIPS↓ &  \multicolumn{1}{|c}{FID↓} & LPIPS↓  & FID↓ & LPIPS↓ & \multicolumn{1}{|c}{FID↓} & LPIPS↓ \\
        \hline
        Latent-Code  & 24.04  & 0.098 & \multicolumn{1}{|c}{26.39} & 0.120 & \multicolumn{1}{|c}{27.48} & 0.125 & 3.21 & 0.097 & \multicolumn{1}{|c}{5.31}  & 0.131 \\
        Pluralistic  & 16.59  & 0.198 & \multicolumn{1}{|c}{17.09} & 0.201  &  \multicolumn{1}{|c}{16.98} & 0.203 & 9.83 & \multicolumn{1}{c|}{0.197}  & 16.26 & 0.226 \\   
        ZITS++  & -  & - & \multicolumn{1}{|c}{-} & - &  \multicolumn{1}{|c}{-} & - & 3.87 & \multicolumn{1}{c|}{0.118}  & 6.13 & 0.376 \\
        RePaint  & 7.26  & 0.066 & \multicolumn{1}{|c}{10.06} & 0.071 & \multicolumn{1}{|c}{9.71} & 0.069 & 11.16 & \multicolumn{1}{c|}{0.102}  & 20.62  & 0.117 \\
        LaMa     & \textcolor{red}{5.26} & \textcolor{blue}{0.037}  & \multicolumn{1}{|c}{\textcolor{red}{8.91}} & \textcolor{blue}{0.057} & \multicolumn{1}{|c}{\textcolor{red}{4.86}} & \textcolor{blue}{0.053} & \textcolor{red}{1.43} & \multicolumn{1}{c|}{\textcolor{blue}{0.061}}  & \textcolor{blue}{3.60} & 0.104 \\
        MAT   & 9.83  & 0.064 & \multicolumn{1}{|c}{9.84} & 0.065 & \multicolumn{1}{|c}{7.15} & 0.068 & \textcolor{blue}{1.51} & \multicolumn{1}{c|}{0.067}  & \textcolor{red}{3.47} & \textcolor{red}{0.087} \\

        \hdashline
        
        Ours   & \textcolor{blue}{6.38}  & \textcolor{red}{0.028} & \multicolumn{1}{|c}{\textcolor{blue}{9.45}} & \textcolor{red}{0.054} & \multicolumn{1}{|c}{\textcolor{blue}{7.03}}  & \textcolor{red}{0.043} & 2.61 & \textcolor{red}{0.056} & \multicolumn{1}{|c}{5.03} & \textcolor{blue}{0.094} \\
        \hline
    \end{tabular}
    \caption{Quantitative evaluation on Places2 and CelebA-HQ datasets. The best and second best results are in \textcolor{red}{red} and \textcolor{blue}{blue}.}    
    \label{tab-celeba}
    
\end{table*}

\begin{figure}
    \centering

    \subfigure[FID \qquad \qquad \qquad \qquad (b)  LPIPS ]{
		\begin{minipage}[t]{1\linewidth}
			\centering
			\includegraphics[width=1\linewidth]{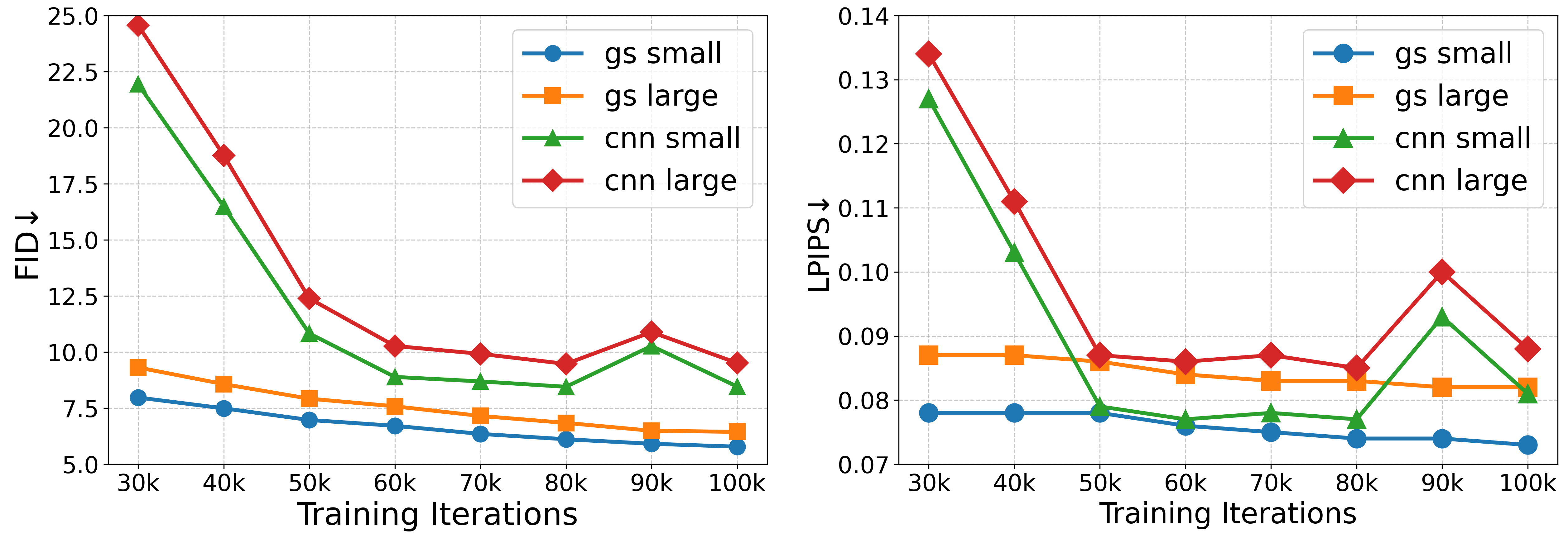}
		\end{minipage}
        \vspace{-1em}
	}
    
    \caption{Convergence Speed comparison between CNN decoder and Gaussian decoder}
    \label{fig-ablation-convergence}
    \vspace{-1em}
\end{figure}

\begin{table}[htbp]
    \centering
    \small
    \begin{tabular}{c|cc|cc}
    \hline

    \multirow{2}{*}{\textbf{Settings}} & \multicolumn{2}{c|}{\textbf{Small}} & \multicolumn{2}{c}{\textbf{Large}} \\
    
    & FID↓ & \multicolumn{1}{c|}{LPIPS↓}  & FID↓ & LPIPS↓   \\
        \hline    
        \rowcolor{pink!30} Full Model & 5.78 &  0.073  & 6.44 & 0.082 \\
        w/o dino map  & 7.34 &  0.076  & 8.37 & 0.085 \\
        w/o dino      & 8.73 &  0.077  & 9.92 & 0.086 \\
        CNN decoder   & 8.46 &  0.081  & 9.51 & 0.088 \\
        \hdashline
        GS-100        & 7.22 &  0.077  & 8.36 & 0.085 \\
        GS-196        & 6.38 &  0.075  & 7.16 & 0.084 \\
        \hdashline
        SA-Adaln-no-$\gamma$  & 7.28 &  0.079  & 8.04 & 0.087 \\
        SA-Concat     & 7.27 &  0.082  & 8.40 & 0.082 \\

    \hline
    \end{tabular}
    \caption{Ablation study results on ImageNet100 dataset.}
    \label{tab-ablation}
    \vspace{-1em}
\end{table}

\subsection{Abalation Study}
This study aims to investigate the impact of our key components through a series of ablation experiments. We adopt ImageNet-100~\cite{deng2009imagenet}, a subset of ImageNet that focuses on natural scenes, due to its balance of efficiency and content diversity. The dataset comprises 130,000 images for training and 5,000 images for testing. For fair comparison, all models are trained for 100,000 steps and evaluated on the same set of image-mask pairs.

\paragraph{DINO Feature}
To assess the effectiveness of the feature alignment module, firstly, we remove the semantic adaptation component and directly use the raw features from DINO. This results in a noticeable degradation in image quality, particularly under large-mask conditions, indicating that unadapted features are insufficient for guiding the inpainting process effectively. Next, we remove the entire semantic alignment module, which leads to a further decline in performance, with more artifacts and structural inconsistencies observed in the inpainted regions as shown in Figure~\ref{fig-ablation-convergence}.

\begin{figure}[!t]
    \centering
    \vspace{-0.5em}

    \subfigure[full model\qquad \quad (b) w/o dino \qquad \protect\ (c) cnn decoder]{
		\begin{minipage}[t]{1\linewidth}
			\centering
			\includegraphics[width=1\linewidth]{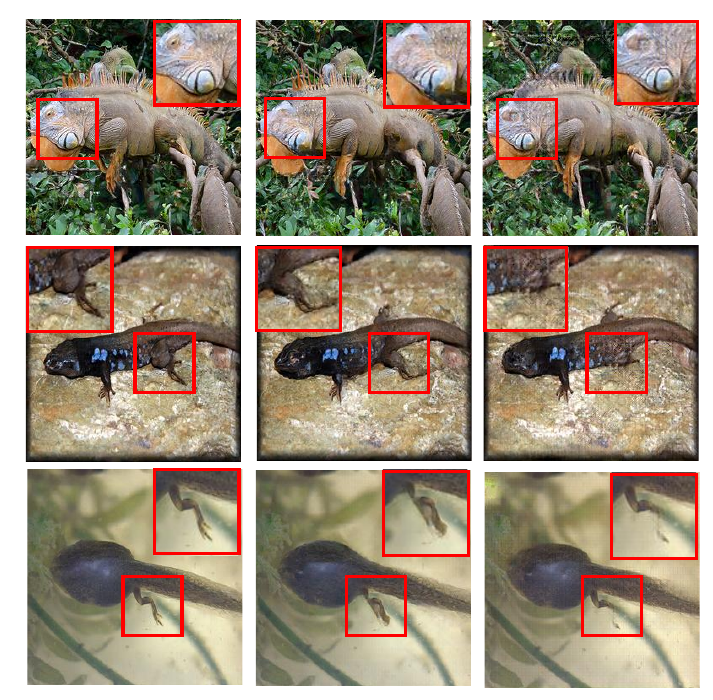}
		\end{minipage}
	}
    \vspace{-1em}

    \caption{{Ablation study}. Visual comparison of inpainting results under different configurations: (a) the full model (b) removing DINO-based semantic guidance (c) replacing our rasterization-based decoder with a CNN-based decoder.}
    \label{fig-ablation}
    \vspace{-1em}
\end{figure}

\paragraph{CNN Decoder}
To assess the rasterization-based decoder, we replace it with a CNN decoder using transposed convolutions on the same patch-level features. Both decoders have comparable parameter counts to ensure fairness. As shown in Table~\ref{tab-ablation}, the CNN decoder suffers a notable performance drop. Figure~\ref{fig-ablation} reveals visual artifacts, while Figure~\ref{fig-ablation-convergence} shows slower convergence and training instability, especially in early stages. In contrast, our Gaussian decoder enables faster, more stable learning with superior results.

\paragraph{Gaussian Numbers}
The number of Gaussians is a critical hyperparameter that significantly affects generation quality. To study its impact, we train models with 100 and 196 Gaussians and compare them against our baseline setting of 324 Gaussians, as shown in Table~\ref{tab-ablation}. As expected, reducing the number of Gaussians leads to a corresponding decline in performance. Intuitively, increasing the number of Gaussians could significantly improve quality by offering finer spatial resolution. However, this significantly raises computational cost and model size, making it impractical for efficient training and inference. We leave this for future work.

\paragraph{Condition Methods}
To assess the effectiveness of our AdaLN module, we conduct two ablation experiments. First, we remove the scaling parameter $\gamma$, resulting in severe training instability. As shown in Table~\ref{tab-ablation}, the reported results are from step 50,000, beyond which training completely collapses. Notably, even before the breakdown, the model underperforms the baseline across several metrics. In the second experiment, we replace AdaLN with a simple concatenation-based fusion while keeping the number of parameters approximately the same to ensure a fair comparison. This variant yields a noticeable drop in performance across all evaluation metrics, further demonstrating the superiority of our AdaLN design.

\section{Conclusion}
In this paper, we present a novel patch-level 2D Gaussian Splatting (2DGS) framework with semantic alignment for image inpainting, effectively addressing the challenge of coherent completion in missing regions. Our approach encodes incomplete images into 2D Gaussian parameters using a lightweight CNN-based U-Net and reconstructs them through a learnable rasterization pipeline. The key contributions include a semantic feature alignment strategy that leverages DINO-based priors to guide the inpainting process, and a patch-level rasterization mechanism that significantly reduces GPU memory usage and improves inference efficiency by operating on localized image blocks. Extensive experiments demonstrate that our method achieves competitive performance, while ablation studies validate the effectiveness of each key component. This work highlights the strong potential of efficient, Gaussian-based representations for realistic image restoration and broader visual synthesis.


\bigskip

\bibliography{aaai2026}

\newpage
\appendix
\urlstyle{rm} 
\def\UrlFont{\rm}  

\frenchspacing  
\setlength{\pdfpagewidth}{8.5in} 
\setlength{\pdfpageheight}{11in} 
%


%
\floatstyle{ruled}
\newfloat{listing}{tb}{lst}{}
\floatname{listing}{Listing}
%
\pdfinfo{
/TemplateVersion (2026.1)
}

\setcounter{secnumdepth}{2} 

%


\title{2D Gaussian Splatting with Semantic Alignment for Image Inpainting \\Appendix}


\appendix
\newpage


\section{Implementation Details}
\subsection{Hyperparameter}
We provide the experiment details on CelebA-HQ and Places2 datasets for reproducibility below:

\begin{table}[htbp]
    \centering
    \begin{tabular}{c|cc}
        \hline
        Settings & CelebA-HQ  & Places2  \\
        \hline
        Image size      &   256   &   256 \\
        Patch size      &   $16\times16$    & $16\times16$  \\
        Gaussian per patch    & 324     &   324   \\
        Hidden dimension per gs      & 12     &  12  \\
        Overlap pad   &  1    &  1   \\
        Batch size    &   64    &   64  \\
        Learning rate   & 2e-4   &   2e-4   \\
        Training steps  & 60k    &   200k   \\
        Optimizer      & Adam    &   Adam   \\
        GAN loss      &  hinge   &  hinge   \\
        D learning rate   &  2e-4    &  2e-4   \\
        D optimizer    &  Adam    &   Adam   \\
        Reconstruction loss weight   & 1   &  1  \\
        Perceptual loss weight   &  2   &   3  \\
        GAN loss weight  &  0.2   &  0.3   \\
        Alignment loss weight   &  1   &   1  \\
         
        \hline
    \end{tabular}
    \caption{Hyperparameters}    
\end{table}

We refer to the small mask as $20\% \sim 40\% $ ratio and the large mask as $ 40\% \sim 60\% $ ratio.

\subsection{Mask Ratio Strategy}
When training the model with varying mask ratios, we progressively increase the masking ratio to help the model adapt more easily. After a certain number of iterations, we begin randomly sampling the ratio from the full range to prevent the model from underperforming on smaller ratios. Concretely, for the CelebA-HQ dataset, we increase the masking ratio every 20 epochs, while for the Places2 dataset, we do so every epoch due to its larger scale.

\section{Social impacts}
\subsection{Positive impacts}
\begin{itemize}
    \item Advancements in AI: A New paradigm enhances the development of a data tokenizer capable of assisting in various tasks.
    \item Practicality: The method can provide assistance for the image restoration field, aiding in future restoration work and helping people repair damaged data.
\end{itemize}

\subsection{Negative impacts}

\begin{itemize}
    \item Job Displacement: Advanced AI could potentially displace jobs in the data restoration field, necessitating consideration of economic and societal impacts.
    \item Misuse: Misuse of our model may lead to data fraud, information infringement, incorrect guidance on the internet, resulting in economic and ethical issues.
\end{itemize}

\section{Discussion}

While our approach achieves strong results in both semantic consistency and visual fidelity, it currently lacks the explicit controllability. These methods often benefit from multi-modal inputs, such as textual prompts or structural cues, that enable more flexible and user-guided generation. In contrast, our design operates without external guidance, which limits its applicability in scenarios requiring fine-grained semantic control or interactive editing. Enhancing our framework with cross-modal conditioning mechanisms is a compelling direction for future exploration.

\section{More Qualitative Results}

\begin{figure*}[!htbp]
    \centering
    \vspace{-1em}
    \includegraphics[width=1.0\linewidth]{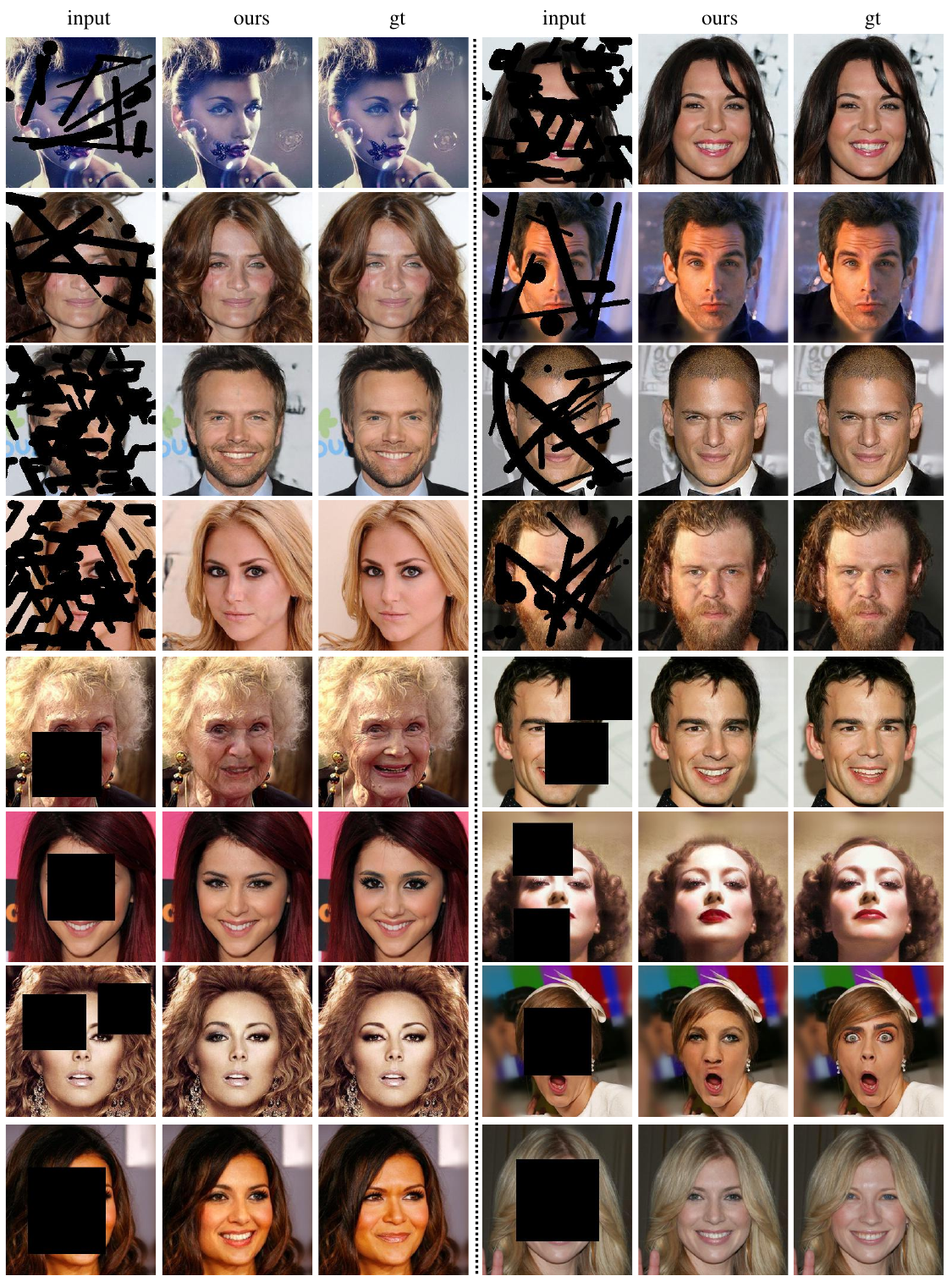}
    \caption{Qualitative results on CelebA-HQ.}
    \label{fig-celeba}
\end{figure*}

\begin{figure*}[!h]
    \centering
    \vspace{-3em}
    \includegraphics[width=1.0\linewidth]{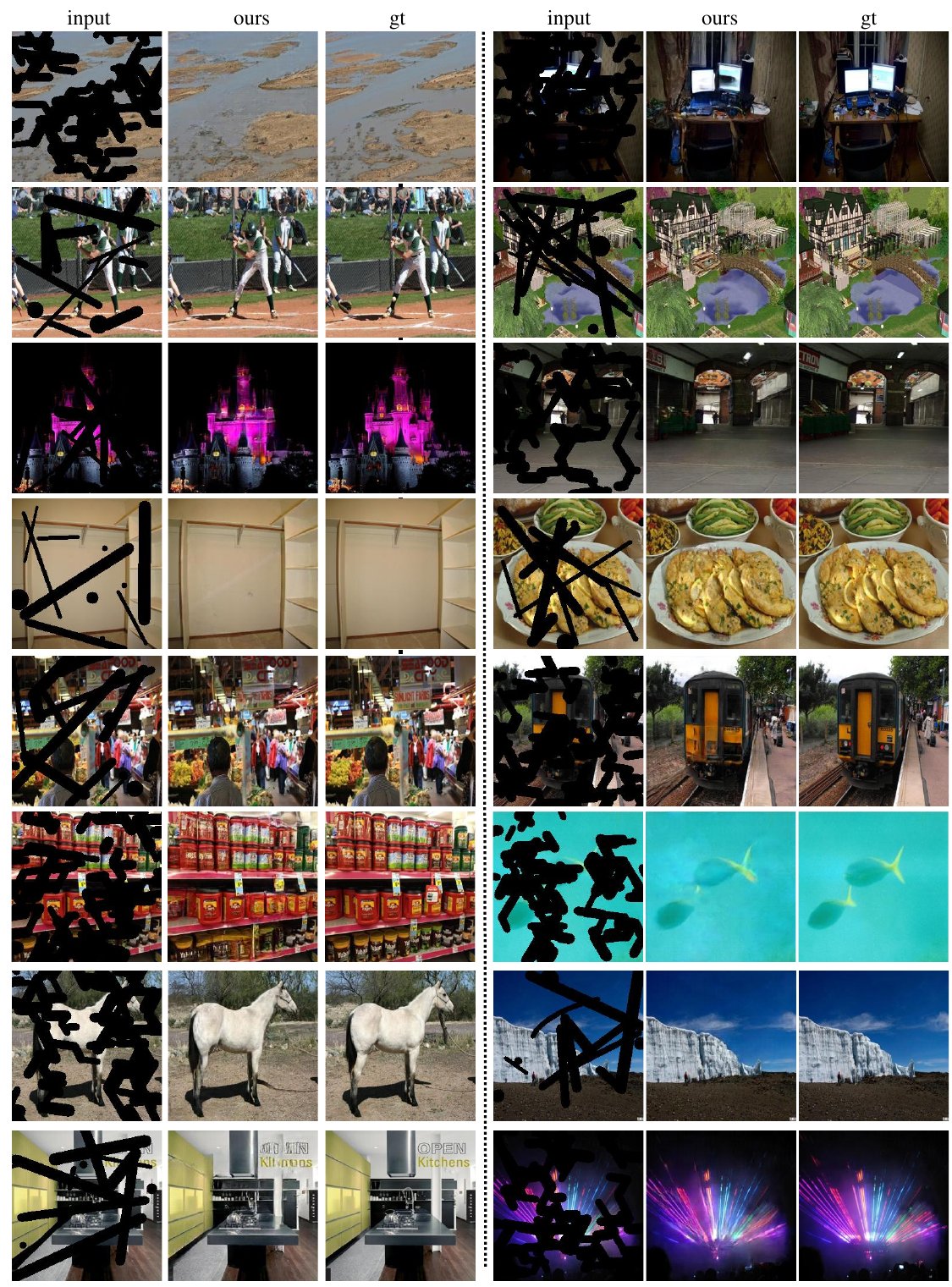}
    \caption{Qualitative results on Places2.}
    \label{fig-places2}
\end{figure*}

\begin{figure*}[!htbp]
    \centering
    \includegraphics[width=0.98\linewidth]{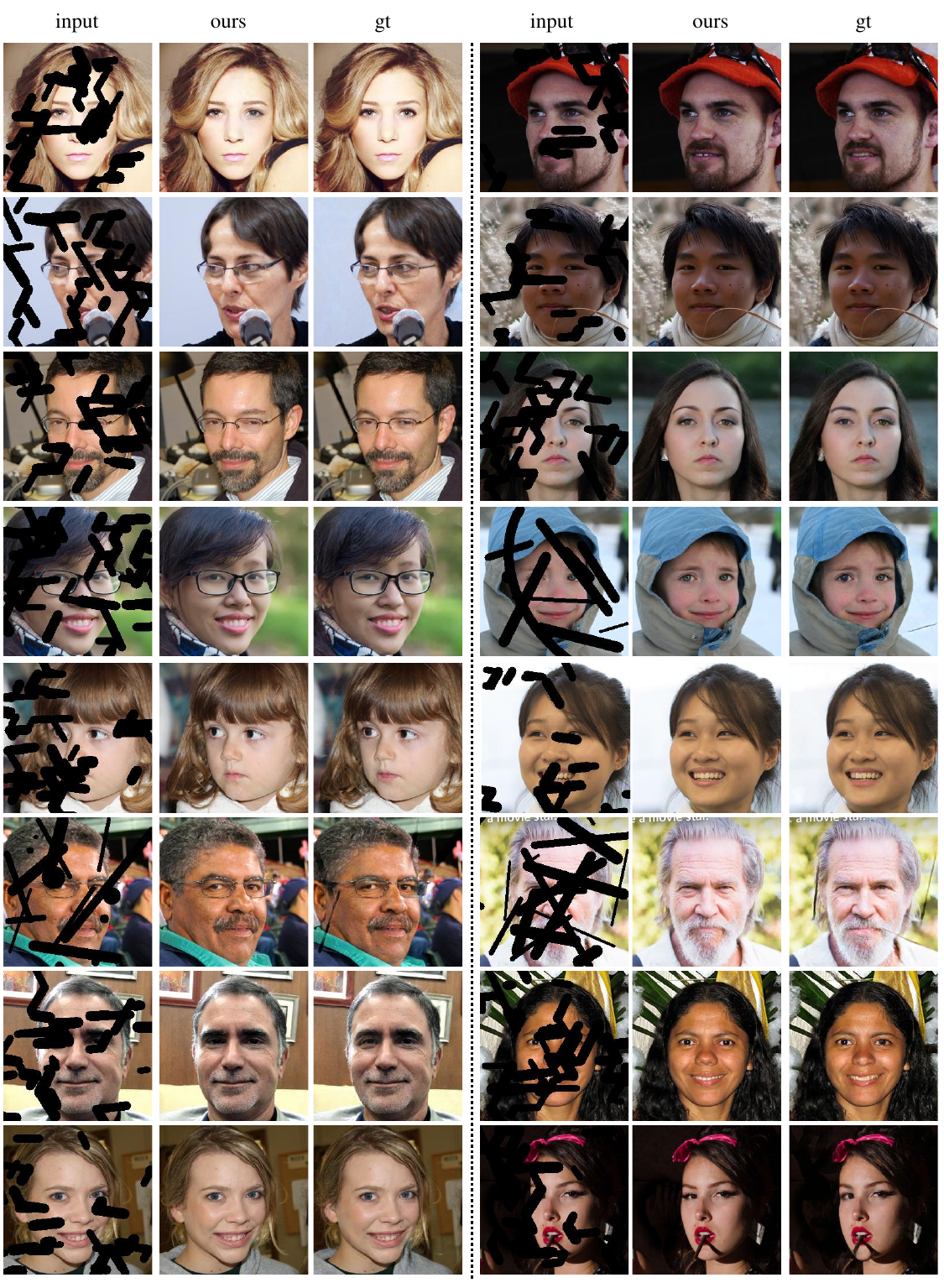}
    \caption{Qualitative results on FFHQ.}
    \label{fig-ffhq}
\end{figure*}

\begin{figure*}[!htbp]
    \centering
    \vspace{-3em}
    \includegraphics[width=1.\linewidth]{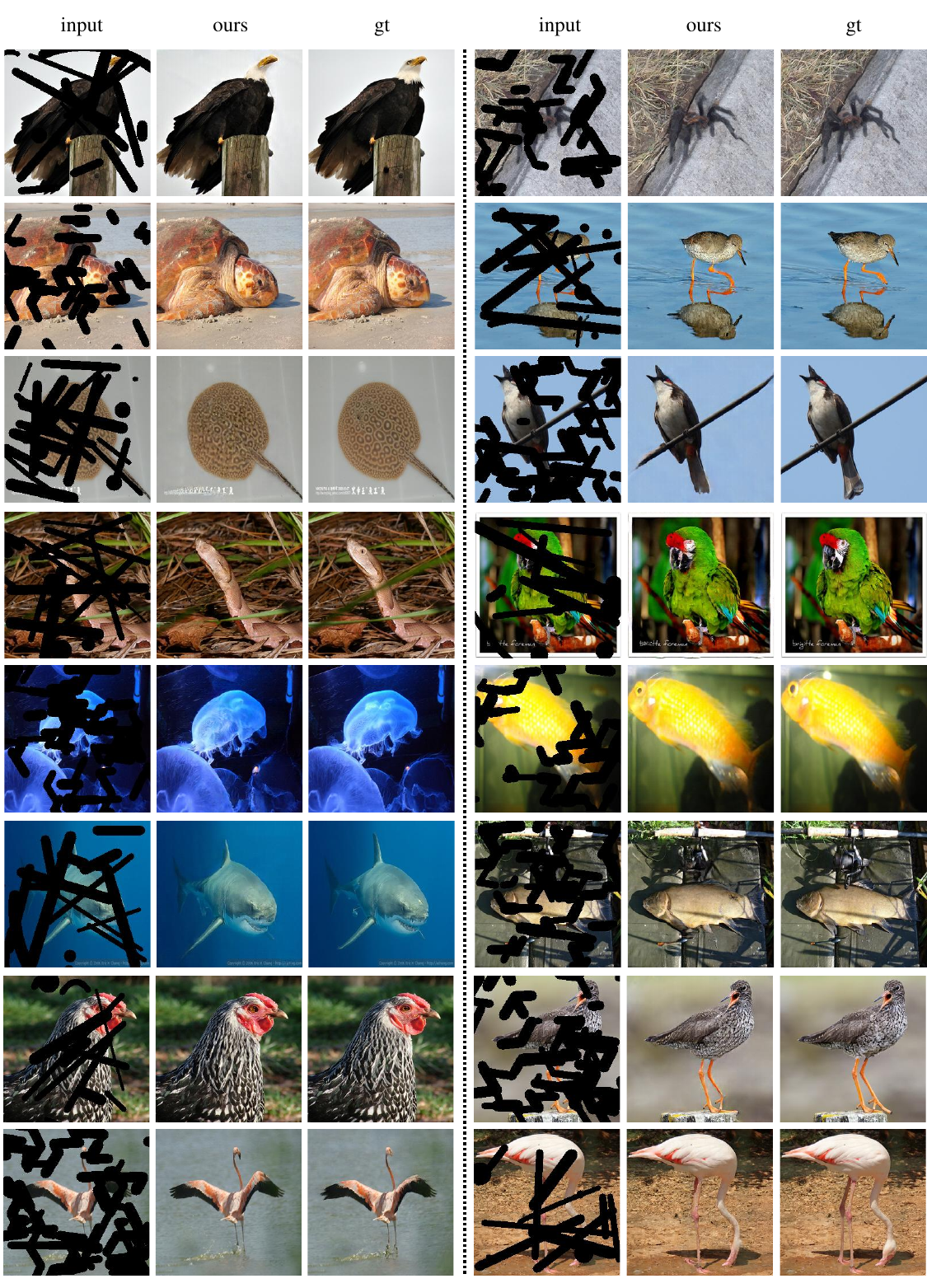}
    \caption{Qualitative results on ImageNet-100.}
    \label{fig-imagenet}
\end{figure*}

To further validate the robustness and generalization of our method, we present qualitative results across diverse datasets and mask settings. As shown in Figures~\ref{fig-celeba}--\ref{fig-imagenet}, our approach consistently produces semantically coherent and visually realistic completions.

On CelebA-HQ (Figure~\ref{fig-celeba}), our model effectively preserves facial structure and identity, even under large, irregular masks. For complex natural scenes in Places2 (Figure~\ref{fig-places2}), it reconstructs rich textures and spatial layouts. Results on FFHQ (Figure~\ref{fig-ffhq}) further highlight its ability to handle high-resolution portraits with fine details. Finally, the performance on the challenging and diverse ImageNet-100 dataset (Figure~\ref{fig-imagenet}) demonstrates strong generalization to object-centric images with varied semantics.

Overall, these results confirm that our method not only maintains semantic fidelity across domains, but also adapts well to arbitrary mask distributions—underscoring its potential for real-world inpainting applications.

\section{Real-world Scenario Inpainting}
\begin{figure}[!h]
    \centering
    \includegraphics[width=1.\linewidth]{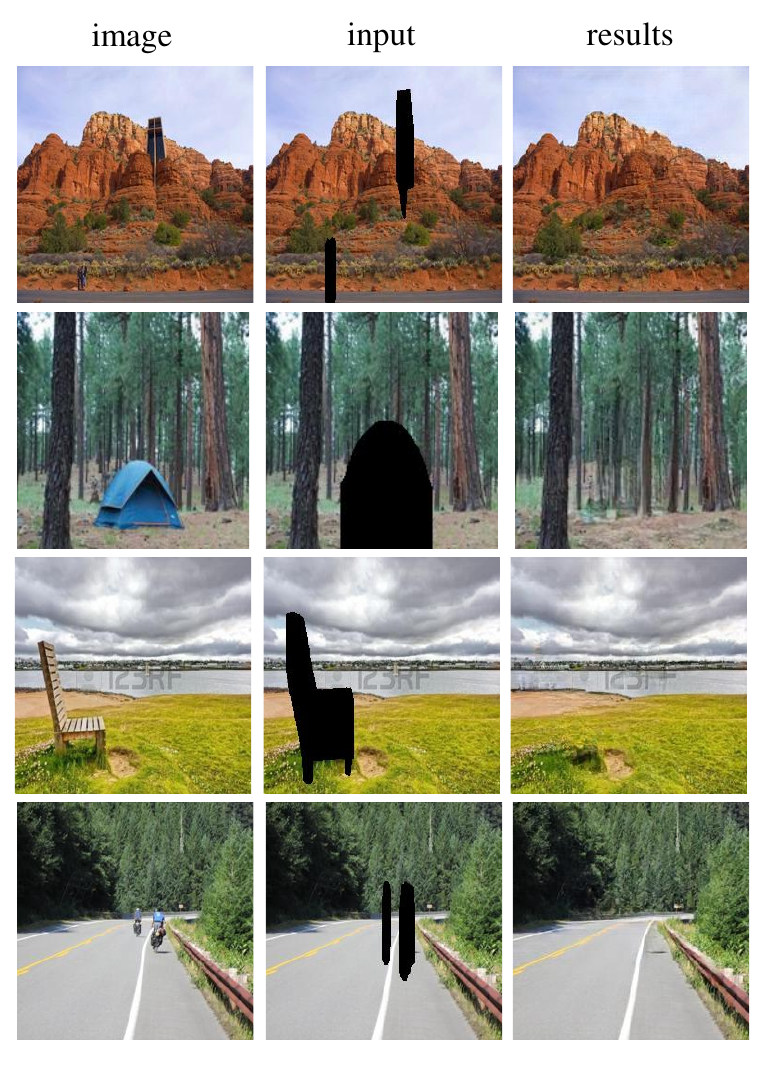}
    \caption{More results on Real-world Scenarios}
    \label{fig-seg}
\end{figure}

As illustrated in Figure~\ref{fig-seg}, we manually remove specific foreground objects from natural images using segmentation masks to simulate realistic inpainting scenarios. This setup mirrors common image editing tasks such as object removal. The masks are designed to selectively occlude frequently encountered elements in outdoor scenes, such as tents, chairs, signage, and people, thereby creating a practical testbed that closely reflects real-world use cases. The second column shows the masked inputs fed into the model, while the final column presents the corresponding inpainted outputs. This visualization not only showcases our model’s ability to reconstruct complex backgrounds with high visual fidelity, but also highlights its effectiveness in preserving structural continuity and semantic coherence in the absence of salient foreground content. Furthermore, the diversity in object shapes and scene compositions underscores the robustness and applicability of our method across a wide range of real-world scenarios.

\newpage
\section{Efficiency}
To assess the computational efficiency of our method, we compare its inference speed with several representative inpainting approaches, as summarized in Table~\ref{tab-speed}. All methods are evaluated under identical hardware settings to ensure a fair comparison. Our approach generates significantly faster than diffusion-based or transformer-heavy methods,

\begin{table}[!h]
    \centering
    \begin{tabular}{c|c}
    \hline
        Method &  Speed(ms) \\
    \hline
       Ours  &  32.52 \\
       LaMa  &  15.80 \\
       MAT   &  65.75 \\
       Latent-Code   &  45.67  \\
      RePaint   &  79035.84   \\
    \hline
    \end{tabular}
    \caption{Average inference time per image.}
    \label{tab-speed}
\end{table}

\end{document}